\documentclass[letterpaper, 10 pt, conference]{ieeeconf} 
\IEEEoverridecommandlockouts  % This command is only needed if you want to use the \thanks command              
\usepackage{amsmath, amssymb}
\usepackage[dvipsnames,table]{xcolor}
\usepackage{siunitx}
\usepackage{stmaryrd}

\usepackage{cite}
\usepackage{booktabs}
\usepackage{xurl}
\usepackage{mwe}
\usepackage{graphics} % for pdf, bitmapped graphics files
\usepackage{multirow}
\usepackage{pifont}
\makeatletter
\let\NAT@parse\undefined
\makeatother
\usepackage{hyperref}

\newcommand{\myparagraph}[1]{\noindent \textbf{#1} ---}

\title{\LARGE \bf
Learning to estimate UAV created turbulence from scene structure observed by onboard cameras}
\author{Quentin Possamaï$^{1}$, Steeven Janny$^{1}$, Madiha Nadri$^{2}$, Laurent Bako$^{3}$ and Christian Wolf$^{4}$% <-this % stops a space
\thanks{$^{1}$Quentin Possamaï and Steeven Janny are with INSA-Lyon, LIRIS UMR CNRS 5205.
{\tt\small quentin.possamai@insa-lyon.fr},
{\tt\small steeven.janny@insa-lyon.fr}}%
\thanks{$^{2}$Madiha Nadri is with Université Claude Bernard Lyon 1, LAGEPP
{\tt\small madiha.nadri-wolf@univ-lyon1.fr}}%
\thanks{$^{3}$Laurent Bako is with Ecole Centrale de Lyon, AMPERE,
{\tt\small laurent.bako@ec-lyon.fr}}%
\thanks{$^{4}$Christian Wolf is with Naver Labs Europe, 
{\tt\small christian.wolf@naverlabs.com}}%
}

\newcommand{\norm}[1]{\left\lVert#1\right\rVert}
\begin{document}

\maketitle
\thispagestyle{empty}
\pagestyle{empty}

% %%%%%%%%%%%%%%%%%%%%%%%%%%%%%%%%%%%%%%%%%%%%%%%%%%%%%%%%%%%%%%%%%%%%%%%%%%%%%%%%
% %%%%%%%%%%%%%%%%%%%%%%%%%%%%%%%%%%%%%%%%%%%%%%%%%%%%%%%%%%%%%%%%%%%%%%%%%%%%%%%%

\begin{abstract}
\noindent
Controlling UAV flights precisely requires a realistic dynamic model and accurate state estimates from onboard sensors like UAV, GPS and visual observations. Obtaining a precise dynamic model is extremely difficult, as important aerodynamic effects are hard to model, in particular ground effect and other turbulences. While machine learning has been used in the past to estimate UAV created turbulence, this was restricted to flat grounds or diffuse in-flight air turbulences, both without taking into account obstacles. In this work we address the complex problem of estimating in-flight turbulences caused by obstacles, in particular the complex structures in cluttered environments. We learn a mapping from control input and images captured by onboard cameras to turbulence. In a large-scale setting, we train a model over a large number of different simulated photo-realistic environments loaded into the Habitat.AI  simulator augmented with a dynamic UAV model and an analytic ground effect model. We transfer the model from simulation to a real environment and evaluate on real UAV flights from the EuRoC-MAV dataset, showing that the model is capable of good \textit{sim2real} generalization performance. The dataset will be made publicly available upon acceptance.
\end{abstract}

% %%%%%%%%%%%%%%%%%%%%%%%%%%%%%%%%%%%%%%%%%%%%%%%%%%%%%%%%%%%%%%%%%%%%%%%%%%%%%%%%
% %%%%%%%%%%%%%%%%%%%%%%%%%%%%%%%%%%%%%%%%%%%%%%%%%%%%%%%%%%%%%%%%%%%%%%%%%%%%%%%%
\section{Introduction}
\noindent
Drones are among the most agile vehicles in robotic platforms used today in industry and research laboratories. Their ability to move quickly in all directions in space is a great opportunity for many robotics applications, but also a difficult task in terms of planning and control. For a UAV to succeed in accomplishing a task, it must have an accurate, stable and robust control law to guarantee satisfactory trajectory tracking. This controller design task is made difficult by the absence of on-board sensors delivering accurate and reliable measurements of the 6D position of the drone.
%In order for a a UAV to achieve demanding tasks, the system must follow receives commands from a controller following a reference. A controller biggest challenge is facing perturbations coming from external phenomenon. A possible solution to improve command performances is to model the perturbation to predict and anticipate them. 

Consequently, the controller relies on noisy measurements of the state of the drone (IMU, GPS and optical flow computed on images captured by an on-board camera are the most frequent sensors) as well as on the identification of an incomplete dynamic model to calculate electrical controls applied to its rotors.
Yet, UAVs are greatly sensible to turbulences such as ground effect or down-wash. Such phenomena are merely impossible to model in real time as they require complex fluid dynamic simulations. The quality of a controller is directly related to the quality of its dynamic model. Naturally, research work has therefore focused on the use of deep learning for the rapid estimation of these turbulences and thus significantly improve the robustness of control. 
Deep neural networks allow to bypass computation constraints while predicting the complex behaviour of ground effect \cite{shi_neural_2019} and in-flight air turbulences \cite{bauersfeld_neurobem_2021}. Nonetheless, existing work relies on marker based motion capture (MoCap), which is one of the most precise pose sensors available, but is expensive and, more importantly, requires complex infrastructure, which restricts usage to experimental platforms. 

\begin{figure}[t] \centering
    \includegraphics[width=0.45\textwidth]{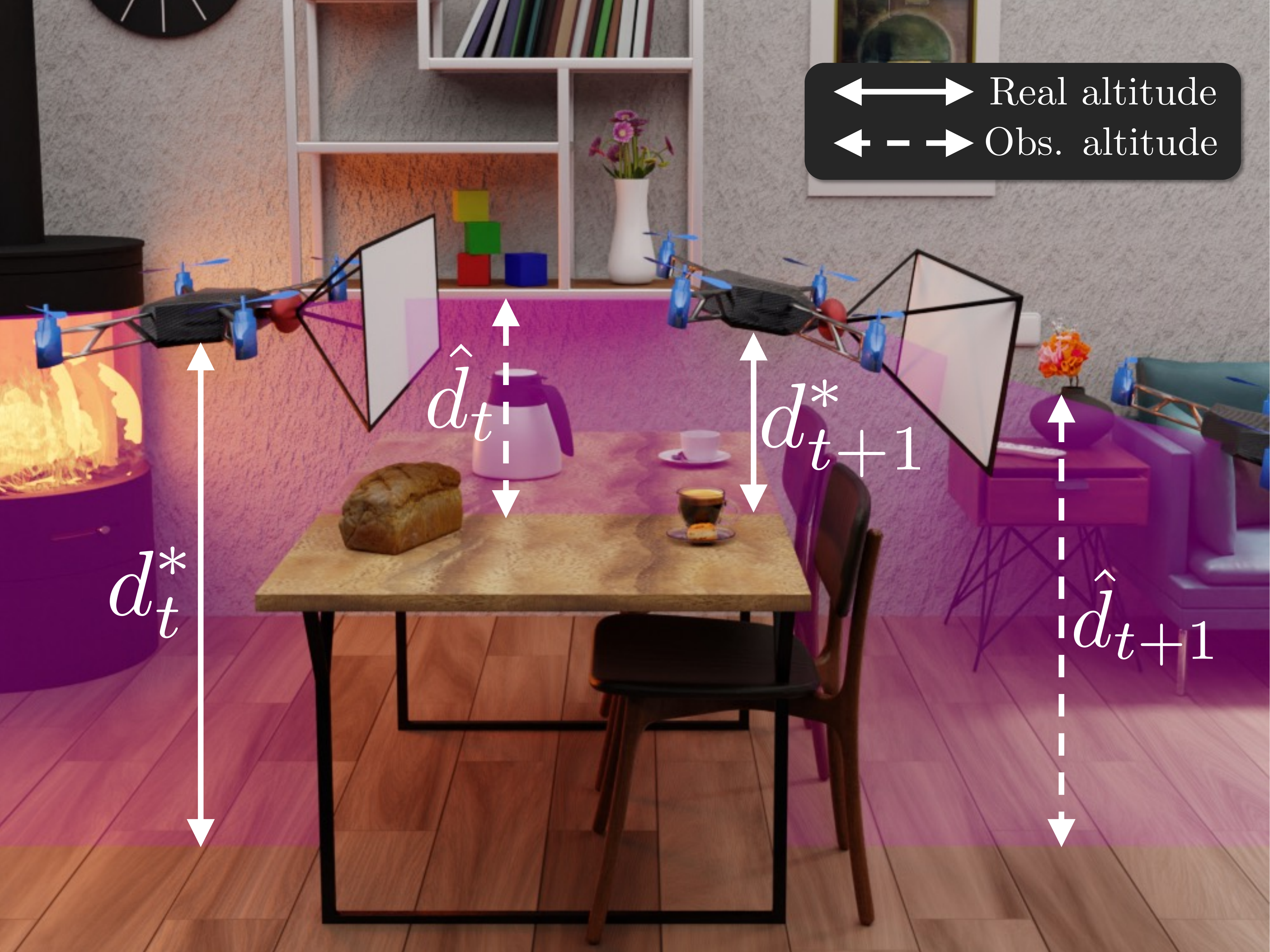}
    \caption{\label{fig:teaser}We estimate turbulences in the form of the ground effect created by UAVs from visual observations captured by onboard cameras. Classical forward oriented cameras create a negative measurement delay when the UAV moves forward, as the currently observed scene structure measures the effect of future turbulences. We deal with this through a memory based model in the form of a recurrent neural network. Our model is trained large-scale in simulation over $\sim$68 different environments of the Matterport 3D dataset \cite{chang_matterport3d_2017} in the Habitat simulator \cite{savva_habitat_2019} and evaluated on real physical drone flights from the EuRoC-MAV dataset\cite{burri_euroc_2016}.}
    \vspace{-5mm}
\end{figure}

In this work, we propose to estimate the turbulences created by the drone itself due to the ground effect. We leverage the causal link between the geometric scene structure observed by the on-board depth camera, and the turbulent air flow. While the ground effect is composed of 3D forces and torques, our work focuses on forecasting solely the vertical force, as it is the most determining effect during hovering flights near surfaces.

Solving this task requires to define a mapping from 2D egocentric images, which are perspective projections of the 3D scene structure, as well as from the control inputs, to turbulences. This mapping should be robust to small irrelevant changes in scene structure and focus on the main factors of variation. It is extremely hard to define this mapping through manual engineering, we therefore propose to leverage the power of high-capacity deep neural networks. The main challenge here resides in the need of large-scale data to learn such a complex mapping, which is extremely difficult to do with real UAV platforms: while it is feasible to collect large amounts of data in terms of equipped UAV flight hours, the need to generalize to new scenes unseen during training also requires diversity in the environments. We need to sample the different factors of variation which are part of the causal chain. In this problem, we need to vary scene structure, i.e. the setup of indoor scenes including positions and type of furniture, positions of walls and doors etc.

Similar to current efforts in robot learning for navigation of terrestrial robots 
\cite{mirowski2017learning,Jaderberg2017Reinforcement, zhu2017target,Chaplot_2020_CVPR,Chaplot2020Learning,beeching2020learning,MarzaArxivMultiON2021}, we propose to leverage 3D photo-realistic simulators like Habitat.AI \cite{savva_habitat_2019} and datasets of 3D scanned indoor environments like Matterport 3D \cite{chang_matterport3d_2017} and Gibson \cite{xia_gibson_2018}. We augment the Habitat simulator with a dynamic drone model and a simulated ground effect following a pre-defined analytical law introduced in \cite{cheeseman_effect_1955}. However, while our method is trained in simulation, we provide transfer to real data and test our model on real UAV flights from the EuRoC-MAV dataset \cite{burri_euroc_2016}.

%Yet, such task is not straighforward : because neural network needs an important quantity of data and real UAV platforms requires engineering and money. 
%Our contribution leverages a procedural data generation proving a ready-to-train dataset, as well as  deep learning model that estimate the surface perturbation of the UAV using an hybrid approach with physics inductive bias. Our model is then tested on real-world data from \cite{burri_euroc_2016}.

This paper is organized as follows: after positioning our contribution with respect to related work in Section \ref{sec:relatedwork}, in Section \ref{sec:dynamicmodelandfa} we describe the dynamic model of the drone and how we use it to compute ground-truth estimates of turbulences from a dataset of real drone flight measurements. Since these data are not abundant enough for training, we use large-scale simulations to generate a large amount of data, described in \ref{sec:simulation}. Section \ref{sec:method} introduces the method for learning to estimate turbulences from on-board sensors, and section \ref{sec:experiments} describes the experiments obtained evaluating on real data after training on simulation.

% %%%%%%%%%%%%%%%%%%%%%%%%%%%%%%%%%%%%%%%%%%%%%%%%%%%%%%%%%%%%%%%%%%%%%%%%%%%%%%%%
% %%%%%%%%%%%%%%%%%%%%%%%%%%%%%%%%%%%%%%%%%%%%%%%%%%%%%%%%%%%%%%%%%%%%%%%%%%%%%%%%

\section{Related Work}
\label{sec:relatedwork}

\myparagraph{Control theory} is classically used to design UAV controllers \cite{invernizzi2018trajectory, marconi2007robust}. Theses approaches are usually based on a physical model assuming extensive knowledge of physical properties such as weight, inertia matrix, or drag coefficient. More accurate dynamic models rely on approximation of aerodynamic laws such as blade element momentum \cite{bauersfeld_neurobem_2021}, or wind field \cite{perozzi_trajectory_2018}. Classical approaches cover a wide range of objectives, ranging from fast and robust stabilization of the attitude, trajectory tracking, path planning and obstacle avoidance \cite{Rubi2020}. Several controller formalisms have been used such as PID (Proportional Derivative Integral), backstepping, sliding mode \cite{moussid_dynamic_2015} \cite{akkinapalli_attitude_2014} or MPC (Model Predictive Control). The latter offers good performances for agile and demanding maneuvers\cite{torrente_data-driven_2021}, yet it requires significant computing resources and a carefully design dynamic model.

\myparagraph{Turbulence modeling} is arguably the most challenging phenomenon to model when dealing with UAVs. Navier-Stokes equations may be used to compute accurate turbulences by modeling the airflow through the rotors \cite{zhang2020numerical,marturano2020numerical}. Yet this solution is intractable for real time control and requires extensive knowledge of the environment. Reduced form of the turbulences, such as ground effect, may be modeled via empirical laws \cite{cheeseman_effect_1955} and learned over a flat surface at take off and landing with a deep neural network using MoCap (marker based Motion Capture) \cite{shi_neural_2019}. In-flight turbulences, including drag and down-wash, are also modeled using hybrid models merging both physics equations and data-driven approaches \cite{bauersfeld_neurobem_2021} and proved to be effective on agile and high speed flights \cite{torrente_data-driven_2021}.

\myparagraph{Deep learning for control} is attracting growing interest from the community. A large body of work is placed at the border of the two disciplines, exploiting the power of data-driven approaches and the guarantees of control theory \cite{janny2021deep,peralez2021Deep,Kolter2019LearningSD}. These approaches have been used for UAV time-optimal trajectory planning \cite{foehn_alphapilot_2020} computer vision for trajectory pursuit \cite{kaufmann_beauty_2019}, or acrobatics maneuvers \cite{kaufmann_deep_2020}.
Moreover, combining control theory and machine learning has shown excellent results. In particular, \cite{torrente_data-driven_2021} and \cite{bauersfeld_neurobem_2021} propose hybrid methods taking advantages of standard controller properties coupled with neural network based state-of-the-art perturbation estimation in the context of UAV control. Associating a physical model with machine learning has also received great interest, since inductive biases are known to facilitate learning \cite{janny2022filteredcophy,de2018deep}. 

%Then \cite{lesort_state_2018} is similar to our work as an encoded current state is learned through observations of state and images in order to predict the next state.

\myparagraph{Robotics and visual navigation}
To cope with the costs of experiments with real robotic equipment and UAVs, reinforcement learning in simulated environments has been proposed as an alternative. Policies can be trained in simulation, allowing the agent to learn complex task without damaging expensive equipment \cite{zengVisualReactionLearning2019,busoniu_reinforcement_2018}. 
The development of photo-realistic 3D simulators \cite{savva_habitat_2019,szot_habitat_2021,kolve_ai2-thor_2019} as well as the release of large-scale datasets of 3D-scanned indoor environment of high diversity \cite{chang_matterport3d_2017,xia_gibson_2018}, and \cite{song_semantic_2017} have enabled the emergence of off-the-shelf technical solutions for terrestrial and aerial navigation \cite{beeching_learning_2020,zengVisualReactionLearning2019}.
The downside of this line of work is the gap between simulation and real environments, which makes the development of production solutions currently difficult. In our work, we show that a model trained in simulation can be deployed to real environments with proper precautions.

%Robot navigation is a huge field in with many challenges to overcome. As learning often requires a lot quantity of data and robotic samples are expensive to obtain, simulating tools \cite{savva_habitat_2019,szot_habitat_2021,kolve_ai2-thor_2019} and realistic datasets \cite{chang_matterport3d_2017,xia_gibson_2018}, and \cite{song_semantic_2017} have been built through time. They offer an important diversity of scenes to represent indoor environments that ease the simulation to reality gap. Flying and terrestrial robots have been navigating using visual input through those environments \cite{beeching_learning_2020,zengVisualReactionLearning2019} and learn to complete task in them.

% %%%%%%%%%%%%%%%%%%%%%%%%%%%%%%%%%%%%%%%%%%%%%%%%%%%%%%%%%%%%%%%%%%%%%%%%%%%%%%%%
% %%%%%%%%%%%%%%%%%%%%%%%%%%%%%%%%%%%%%%%%%%%%%%%%%%%%%%%%%%%%%%%%%%%%%%%%%%%%%%%%

\section{Estimating Ground-Truth turbulences}
\label{sec:dynamicmodelandfa}
\noindent
We focus on UAV flights in complex and cluttered  environments, where the ground effect is an important force disturbing the system. In order to estimate these turbulences on real world flight data, we propose a partial dynamical model, where the propellers thrust and torque are explicitly taken into account. By comparing the ground truth UAV state with the prediction from the model, we extrapolate perturbation forces from the residual error, which we define as the ground effect. This is of course an approximation, as there might be other unmodeled effects, but we assume other sources of perturbation to be negligible compared to ground effects. % but we will ignore these in this work. %In this section we will describe the analytical dynamic model of the drone and the estimation of the ground effects as a residual forces with respect to this model.

\subsection{Dynamical drone model}
\label{par:drone_model}
\noindent
To model the physics of the drone during simulation and to compute the residual for real flights, we adapted a drone model introduced in \cite{Furrer2016} and identified its parameters.
The model has seven parameters, $m$ the mass of the system, $\mathbf{J} = \text{diag}(J_x, J_y, J_z)$ the inertia matrix, $l$ the length of the rotor arms, $k_T$ and $k_M$ respectively the thrust and torque coefficients. We assume $J_x = J_y$. With $\mathcal{F}$ the fixed reference frame, the state $\mathbf{x}$ of the system is composed of $\mathbf{x} = [p, q, v_\mathcal{F}, \omega_\mathcal{B}]$ with $p$ the position and $q$ the quaternion of the UAV's body $\mathcal{B}$ w.r.t $\mathcal{F}$, $v_\mathcal{F}$ the linear velocity expressed in $\mathcal{F}$ and $\omega_\mathcal{B}$ the angular velocity expressed in $\mathcal{B}$. 

\par The propellers produce a vertical thrust in $\mathcal{B}$ noted $\mathbf{T}_u$. The difference in rotational speeds also produces torque $\boldsymbol{\tau}_u$. Letting $\mathbf{\Omega}$ be the vector of rotation speeds of each rotors, one can calculate the resulting forces using the coefficients $k_T$ and $k_M$ and $\mathbf{M}$, a ($4\times N)$ matrix (with $N$ the number of propellers) that depends on the geometrical configuration of the drone:
\vspace{-2mm}
\begin{equation}
    \left[\begin{array}{cc}
         T_u & \boldsymbol{\tau}_u
    \end{array}\right]^T = \left[\begin{array}{cccc}
         k_T &
         lk_T &
         lk_T &
         k_M
    \end{array}\right]^T \odot \mathbf{M}  \mathbf{\Omega}^2
\end{equation}
where $\odot$ is the Hadamard product. The dynamic model is then given by:
\begin{eqnarray}\label{eq_model}
\begin{array}{cl}
    \dot{\mathbf{p}} & = \mathbf{v}_{\mathcal{F}} \\
    \dot{\mathbf{q}} & = \frac{1}{2} \mathbf{q} \circ \boldsymbol{\omega}_\mathcal{B} \\
    \dot{\mathbf{v}}_\mathcal{F} & = \frac{1}{m} \mathbf{q} \circ \mathbf{T}_u \circ \bar{\mathbf{q}} + \mathbf{f}_a - \mathbf{g}\\
    \dot{\boldsymbol{\omega}}_\mathcal{B} & = \mathbf{J}^{-1} ( - \boldsymbol{\omega}_\mathcal{B} \times \mathbf{J} \boldsymbol{\omega}_\mathcal{B} + \boldsymbol{\tau}_u + \bar{\mathbf{q}} \circ \boldsymbol{\tau}_a \circ \mathbf{q})
\end{array}
\label{eq:dynamicmodel}
\end{eqnarray}
with $\circ$ the quaternion product, $\times$ the cross product, $\mathbf{g}$ the gravity acceleration and $\mathbf{T}_u = \left[1\;1\;T_u\right]^T$. $ \bar{\mathbf{q}}$ is the conjugate of  $\mathbf{q}$. $\mathbf{f}_a$ and $\boldsymbol{\tau}_a$ designate the ground effect disturbances that are to be determined. To ease reading of what follows, we will refer as $\mathbf{f}_a$ for both forces and torques due to turbulences.
System (\ref{eq_model}) can be written in matrix form as:
\begin{equation}\label{eq_mosel_matrix}
    \dot{\mathbf{x}}=f(\mathbf{x},\boldsymbol{\Omega},\mathbf{f}_a),
\end{equation}
%where $x$ denotes the state, $\mathbf{I}$ is image from the camera on the drone. 
In the rest of this paper, we will consider a discrete time version of \eqref{eq_mosel_matrix}. System parameter identification is performed using the ground truth state measurements that may be acquired using motion capture or accurate visual odometry algorithm \cite{geneva_openvins_2020,qin_vins-mono_2018}. The dynamical system can be written in a linear form with regards to four parameters :
\begin{multline}
    \left[\begin{array}{cc}
         \dot{\mathbf{v}}_\mathcal{F}  \\
         \dot{\boldsymbol{\omega}}_\mathcal{B}
    \end{array}\right] = \mathbf{A} \left[\theta_0 \; \theta_1 \; \theta_2 \; \theta_3\right]^T + \mathbf{b} \text{ with } \mathbf{b} = \left[\begin{array}{c}
         -\mathbf{g}  \\
         -\omega_y\omega_z \\
         \omega_x\omega_z \\
         0
    \end{array}\right] \\
    \mathbf{A} = \left[\begin{array}{cccc}
         \mathbf{R}^{-1}(\mathbf{M}\boldsymbol{\Omega})_1 & 0 & 0 & 0  \\
         0 & (\mathbf{M}\boldsymbol{\Omega})_2 & 0 & \omega_y\omega_z\\
         0 & (\mathbf{M}\boldsymbol{\Omega})_3 & 0 & \omega_x\omega_z\\
         0 & 0 & (\mathbf{M}\boldsymbol{\Omega})_4 & \omega_y\omega_z\\
    \end{array}\right]
\end{multline}

\noindent
with $\theta_0 = \frac{k_T}{m}$, $\theta_1 = \frac{lk_T}{J_x}$, $\theta_2 = \frac{k_M}{J_z}$, $\theta_3=\frac{J_z}{J_x}$ and $\mathbf{R}$ is the rotation matrix. Under this form, parameter identification can be readily conducted using linear least square optimization.

\subsection{Estimating GT turbulences}
\label{subsec:estimating_gt_turbulence}
\noindent
We estimate ground-truth (GT) values for the turbulences $\mathbf{f}_a$ through residual forces with respect to the forces predicted of the dynamical model described above. This is done with a drone equipped with marker based motion capture (MoCap), which provides state estimates for each time instant. Note, that this is done to obtain GT values only, this method is \textit{not} applicable for turbulence estimation on the fly. Our proposed method for online estimation from visual observations is described in section \ref{sec:method}.

We estimate GT turbulences on real drone flight data from the EuRoC-MAV dataset \cite{burri_euroc_2016}. We focus on the hexarotor UAV flights in the V1 room. The dataset provides state estimates $\{\mathbf{x}_t\}_{t=1..T}$ from motion capture, motor commands $\{\boldsymbol{\Omega}_t\}_{t=1..T}$ and stereo images from onboard camera.
%, $D_{\text{euroc}} = (\hat{\mathbf{x}}_n, \boldsymbol{\Omega}_n, \mathbf{I}_n)_{n \in [\![0, N-1]\!]}$ where $\hat{\mathbf{x}}_n$ is an estimated of a part of the state, $\mathbf{u}_n$ is the command, $\mathbf{I}_n$ is one image or two stereo images from the camera on the drone for the time-step $n$.
%The turbulence forces $\mathbf{f}_a$ will be estimated with a neural network $\mathbf{f}_{a,n}(\hat{\mathbf{x}}_n, \mathbf{u}_n, \mathbf{I}_n; \bm\Theta_a)$.
We estimate GT turbulences $\mathbf{f}^*_a$ as a residual with respect to the identified dynamic drone model, 
\begin{equation}
\mathbf{f}^*_{a,t} = \arg \min_{\mathbf{f}_{a,t}}  \norm{\mathbf{x}_{t+1} - \left(\mathbf{x}_{t} + \text{d}t f({\mathbf{x}}_t, \boldsymbol{\Omega}_t, \mathbf{f}_{a,t}) \right)}^2 
\end{equation}
\noindent with $\text{d}t$ the sampling time of the MoCap signal.

\section{Large-scale generation of simulated UAV trajectory data}
\label{sec:simulation}
\noindent
In the previous section we described a method to estimate turbulence data from UAVs equipped with MoCap, which is not applicable in realistic situations. In this section we will describe a simulated environment capable of generating large-scale amounts of simulated data, which allows us to train a neural model capable of turbulence prediction from visual data alone in realistic situations, and which will then be introduced in section \ref{sec:method}.

The simulated environment is composed of two parts: photo-realistic simulations of observed camera images taken from a virtual UAV flying through virtual environments, and an analytical ground effect model.

\subsection{Photo-realistic visual data}
\begin{figure*}[t]
    \centering
    \includegraphics[width=\linewidth]{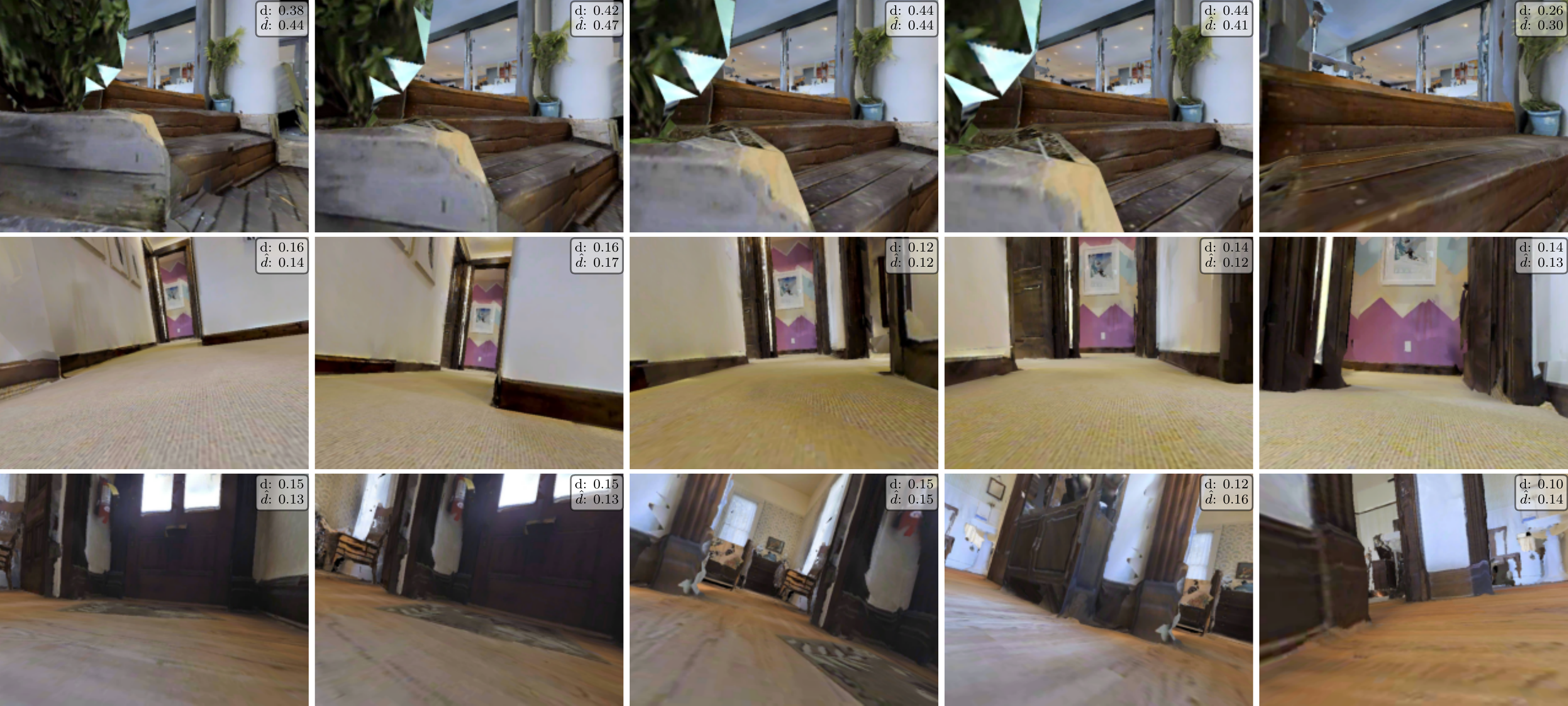}
    \caption{Illustration of estimation results over several simulated example trajectories. We show observed frames, ground truth height-over-ground and estimates predicted by the neural model.}
    \label{fig:habitat_trajectories}
    \vspace{-5mm}
\end{figure*}

\noindent
We use the Habitat.ai \cite{savva_habitat_2019} simulator and load the Matterport 3D dataset \cite{chang_matterport3d_2017}, which consists of 90 different indoor environments in the form of textured meshes produced by 3D scanned apartments, houses and buildings. The simulator is capable of moving a virtual agent around the scene and produce simulated camera images --- see Figure \ref{fig:habitat_trajectories} for examples. The simulator contains a very generic physics engine capable of calculating collisions.

As Habitat has been designed and optimized for terrestrial robots, we augmented it with the UAV dynamic model given in equation (\ref{eq:dynamicmodel}), section \ref{sec:dynamicmodelandfa}. We generate realistic drone flight trajectories through a model predictive controller (MPC) following reference trajectories, which we sample randomly following strong constraints, as follows. (i) A navigation mesh is extracted from the environment mesh, i.e. a 2D mesh which indicates where agents can navigate, taking account of walls and other obstacles. (ii) Data points are randomly sampled on this mesh, and (iii) ordered as a trajectory by solving a travelling salesman problem with the Christofides algorithm \cite{Christofides2015}. We leverage the navigation mesh to compute additional points in order to prevent crossing walls. Finally, the reference is a sequence of straight lines connecting these points with its yaw facing forward. 
Collision checking is implemented, as the MPC controller cannot follow the reference trajectory precisely given its acute angles. %To reduce the chance of colliding, before simulating, a larger cube moves through the reference and modify according to the collision. This cube represents the majority of the controller's error. 

% \begin{figure}[t] \centering
%     \includegraphics[width=0.8\linewidth]{graphics/navmesh.jpg}
%     \caption{\label{fig:reftraj}Illustration of a reference trajectory produced for an indoor scene of the Matterport 3D dataset \cite{chang_matterport3d_2017}: points are randomly sampled at the NavMesh and ordered by solving a travelling salesman problem.}
% \end{figure}

The full dataset is constituted of $531$ trajectories along $68$ scenes and a total of $358,828$ samples. The scenes are split into 43 for training, 9 for validation and 16 for testing according to \cite{chang_matterport3d_2017}. A trajectory is composed of samples $(\mathbf{x}, \boldsymbol{\Omega}, \mathbf{I}, \mathbf{D}, d, \mathbf{f}_a)$ with $\mathbf{x}$ the state of the drone described above, $\boldsymbol{\Omega}$ the rotors rotation speed, $\mathbf{I}$ the image captured by the UAV's camera, $\mathbf{D}$ the depth-map associated to the image, $d$ the height of the UAV w.r.t. the surface below it, and $\mathbf{f}_a$ the ground effect generated by the law given in the next sub-section.

\subsection{Ground-effect model}
\label{sec:cheeseman}
\begin{figure}[t]
    \centering
    \includegraphics[width=0.98\columnwidth]{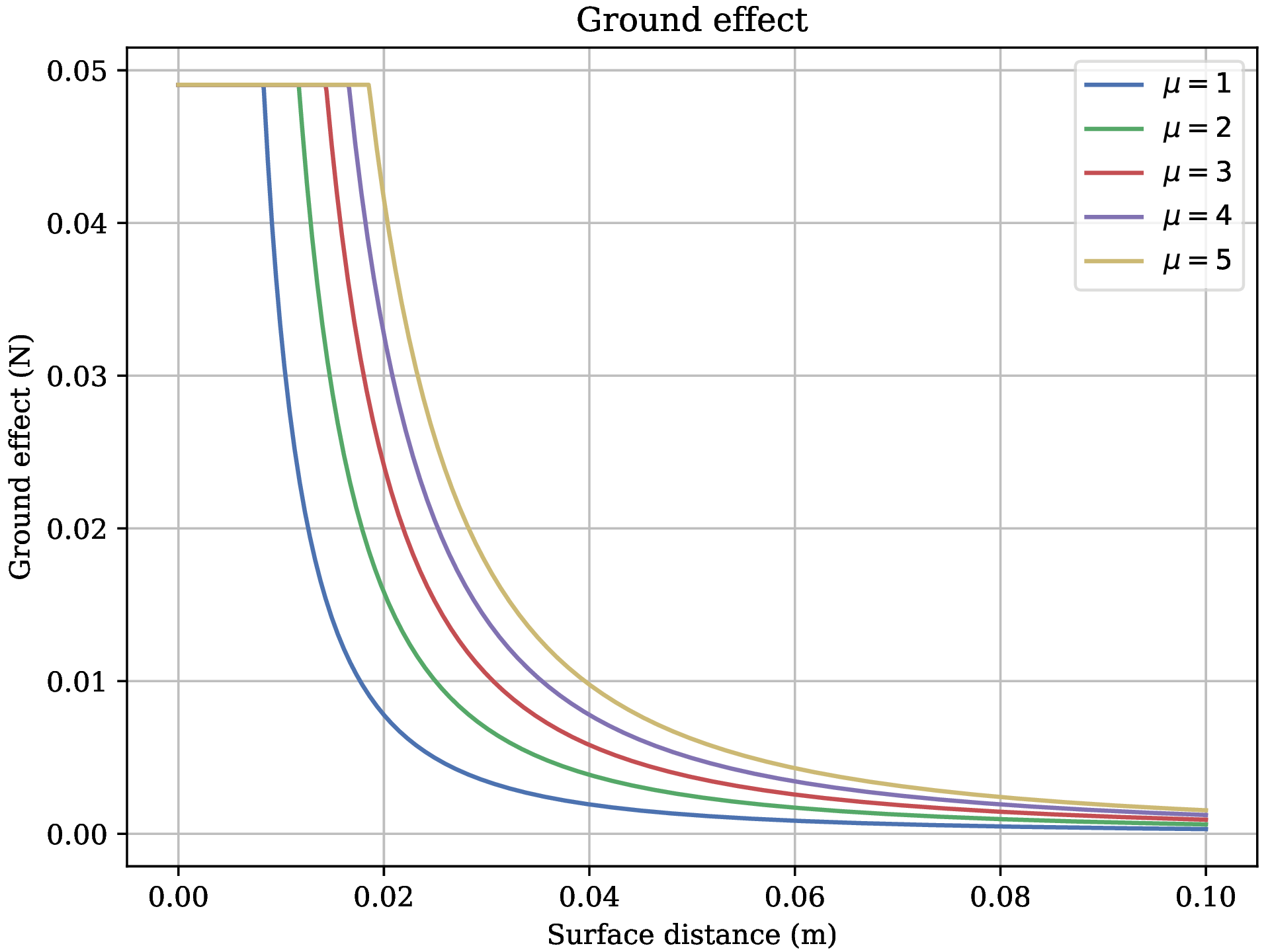}
    \caption{Ground effect used to in the simulation affecting the drone when flying above surfaces. Corresponding to a drone of mass $\SI{50}{\gram}$. $\Omega$ such has the drone is in an hovering state (counterbalance his weight) (i.e $\Omega = [150, 150, 150, 150])$ rad/s, $k_T = 5.45\times 10^{-6}$ SI units, $r = \SI{0.01}{\meter}$.}
    \label{fig:ground_effect}
    \vspace{-5mm}
\end{figure}

\noindent
We simulate the ground effect through an analytical model, which predicts the disturbance given rotor rotation speed and the height of the UAV over ground. This model is certainly an approximation compared to a real simulation of the fluid mechanics involved in the flight; however, as we will see in the experimental section, this law provides quite accurate estimates of the real turbulence values $f^*_a$ estimated with the procedure in section \ref{sec:dynamicmodelandfa}.
Similar to \cite{shi_neural_2019}, we model the ground effect with the law originally introduced in  \cite{cheeseman_effect_1955} and which provides a ground effect estimate affecting the UAV only with a vertical force $f_{a}$:
\begin{equation}\label{eq_ground_effect}
f_{a} = \phi(d, \Omega) = \frac{\sum_i \Omega_i^2 k_T \mu \left(\frac{r}{4 d}\right)^2}{1 - \mu \left(\frac{r}{4 d}\right)^2} 
\end{equation}
%\begin{eqnarray}\label{eq_ground_effect}
%f_{a} = &\phi(d, \Omega) = \frac{\sum_i \Omega_i^2 k_T \mu \left(\frac{r}{4 d}\right)^2}{1 - \mu \left(\frac{r}{4 d}\right)^2} \text{ if } > \sqrt{\mu} r / 4\\
%f_{a} = &\SI{0.1}{\milli \gram} \text{ if } d<\sqrt{\mu} r / 4 \nonumber
%\end{eqnarray}
where $f_{a}$ is the ground effect,  $d$ the surface distance to the UAV and $\Omega_i$ the $i^{th}$ rotor rotation speed. Parameters of the law are $r$ the propeller radius  and $\mu$ a coefficient depending on the number of rotors.  We clip $f_{a}$ to $\SI{0.1}{\milli \gram}$ when $d$ reach values under  $ r\sqrt{\mu}  / 4$. \\
Figure \ref{fig:ground_effect} visualizes the ground effect law as a function of the height over ground $d$ for different value of $\mu$, constant rotor speed and parameters.
%As described in figure \ref{fig:ground_effect}, this disturbance depends on the rotor speed of the UAV $n$ and on the surface under the robot. Then the information to predict the ground effect relies on the surrounding structure of the scene.

% %%%%%%%%%%%%%%%%%%%%%%%%%%%%%%%%%%%%%%%%%%%%%%%%%%%%%%%%%%%%%%%%%%%%%%%%%%%%%%%%
% %%%%%%%%%%%%%%%%%%%%%%%%%%%%%%%%%%%%%%%%%%%%%%%%%%%%%%%%%%%%%%%%%%%%%%%%%%%%%%%%
\section{Learning to estimate Turbulence}
\label{sec:method}
\noindent
In this section we will describe the main contribution consisting of a method to estimate the ground effect from images captured with a forward facing camera. This method does \textit{not} require the marker based MoCap equipment used in section \ref{sec:dynamicmodelandfa} to estimate GT turbulences, it is applicable to any realistic UAV equipment for both indoor and outdoor environments. We propose a hybrid model, which decomposes turbulence prediction into two different phases, as illustrated in Figure \ref{fig:neural_architecture}b: (i) a trained model predicts the height over ground $d_t$ for instant $t$ given an observed depth image $\mathbf{D}_t$; (ii) height is translated into turbulence predictions with the analytical ground effect model introduced in Section \ref{sec:cheeseman}, with parameters properly identified.

\subsection{Learning to estimate height over ground}
\noindent
Height over ground at time $t$ is predicted taking as input the depth image $\mathbf{D}_t$ observed at time $t$. To address the negative measurement delay due to the forward facing direction of the camera (if the drone moves forward, the currently observed scene structure may measure the effect of future turbulences), different instants are not dealt with independently, but rather modelled with a recurrent memory that gathers information from $H$ previous observations. Prediction at time $t$ is given by: %This can be written as follows,
\begin{align}
\label{eq:cheeseman}
\mathbf{e}_{t'} = \textrm{Enc} (\mathbf{D}_{t'}; \theta_\text{Enc}) &   \\
\mathbf{h}_{t'} = \textrm{GRU} (\mathbf{e}_{t'}, \mathbf{h}_{t'-1}; \theta_{\text{GRU}}) &\text{ for } t'\in \llbracket t-H+1, t\rrbracket \label{eq:gru}\\
d_t = \textrm{Pred} (\mathbf{h}_t; \theta_{\text{Pred}})& \label{eq:pred} 
\end{align}
where $\textrm{Enc}()$ is a convolutional encoder network which encodes depth images $\mathbf{D}_t$ into embeddings $\mathbf{e}_t$; $\textrm{GRU}()$ is a recurrent gated unit with recurrent state $\mathbf{h}_t$  and trainable parameters $\theta_{\text{GRU}}$. In this simplified notation we omitted equations of gates and state updates. $\textrm{Pred}()$ is a network which predicts the height over ground $d_t$ from the enriched embedding $\mathbf{h}_t$. Precise architectures for all networks will be given in the experimental section. We supervise this prediction from ground truth heights $d^*$ available from the simulated data described in section \ref{sec:simulation} using an $L_2$ loss $\mathcal{L}_1 = || d_t - d^*_t ||^2$.

\begin{figure}[t] \centering
    \includegraphics[width=\columnwidth]{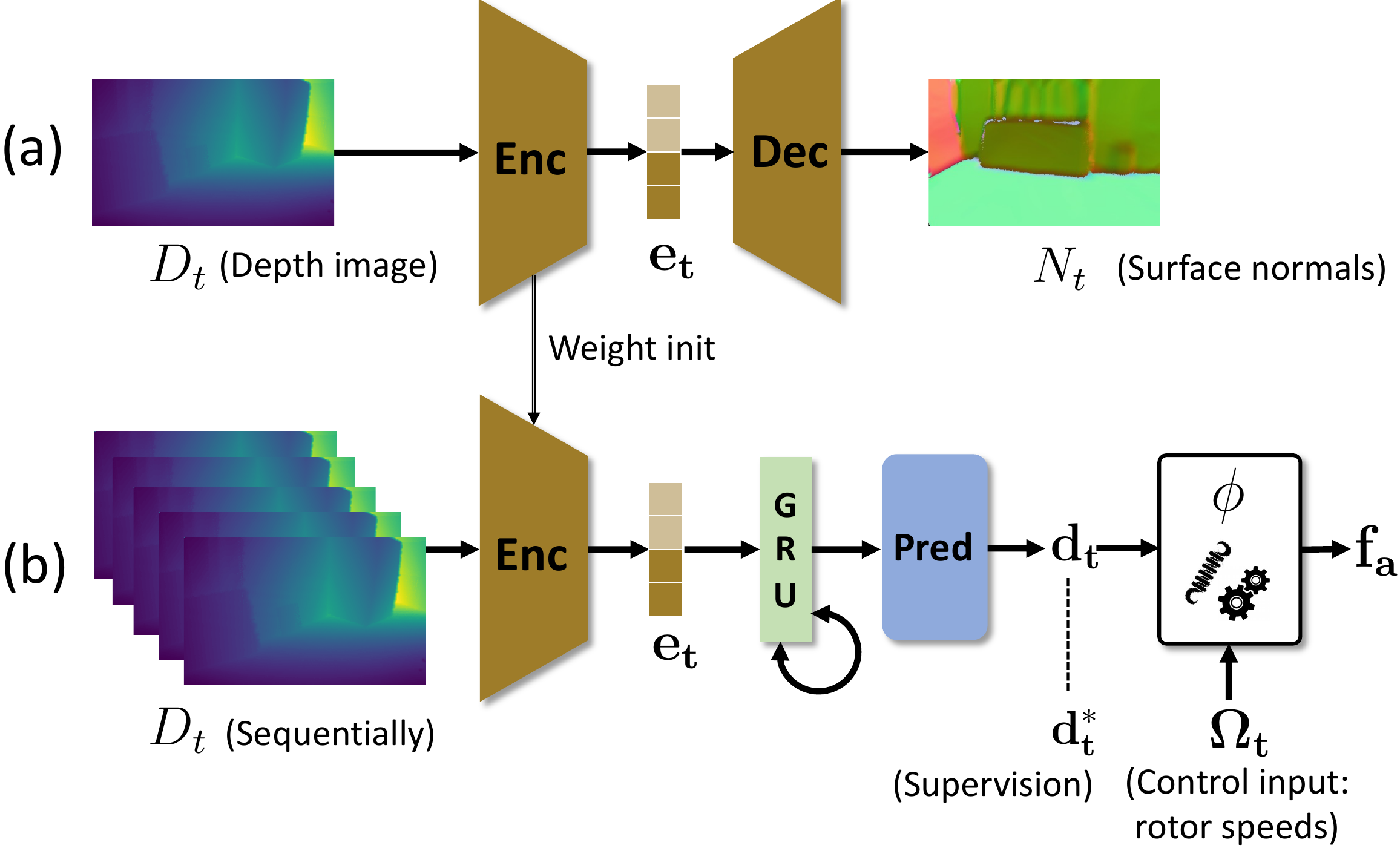}
    \caption{\label{fig:neural_architecture}The neural network architecture. (a) A perception module takes depth inputs $\mathbf{D}_t$ and is pre-trained to predict surface normals $\mathbf{N}_t$ of the same scene. (b) For the final model, the pre-trained encoder is fine-tuned and applied to each observation of the trajectory. Temporal effects are modelled with a gated recurrent unit (GRU), which predicts height over ground $d$. The analytical ground effect model $\phi$ estimates turbulences $f_a$.}
    \vspace{-5mm}
\end{figure}

Training a high-capacity neural network like the encoder function Enc$()$ from a sparse supervision signal like the loss on height alone is challenging. We propose an additional pre-training of the encoder parameters $\theta_{\text{Enc}}$ from an additional data source. As the estimation of height over ground is naturally linked to reconstruction of the 3D scene structure, we pre-train the encoder in an auto-encoding-like transcoder setting, where each depth image $\mathbf{D}_t$ is transcoded into an image $\mathbf{N}_t$ containing the surface normals of the same scene, illustrated in Figure \ref{fig:neural_architecture}a. This creates dense supervision of scene structure with a loss traceable to each pixel in the image. To this end, an additional decoder is added:
\begin{align}
    \mathbf{e}_t = \textrm{Enc} (\mathbf{D}_t; \theta_{\text{Enc}})   \\
    \mathbf{N}_t = \textrm{Dec} (\mathbf{e}_t; \theta_{\text{Dec}})  
\end{align}
We minimize a transcoding loss supervised with the ground truth normals $\mathbf{N}^*_t$
$$
\mathcal{L}_2 = 
\lambda 
|| \mathbf{N}_t - \mathbf{N}^*_t ||^2
+
\sum_{c=1}^3 || \nabla \mathbf{N}_{t,c} - \nabla \mathbf{N}^*_{t,c} ||^2
$$
where $\mathbf{N}_t=\{\mathbf{N}_{t,1}, \mathbf{N}_{t,2}, \mathbf{N}_{t,3}\}$ is a three channel image, with each pixel containing the three coordinates of the surface normals at the given 3D scene point, and $\nabla \mathbf{N}_{t,c}$ is the spatial gradient calculated over channel $c$. The second loss terms adds additional regularization and enforces precise reconstruction in important high-frequency areas of the scene, $\lambda$ is a weighting parameter.

We perform this pre-training on 22,705 images of the IRS dataset \cite{wang_irs_2019}, which contains pairs of synthetic indoor depth and normal images. After pre-training, the decoder is removed and replaced with the pipeline predicting height $d_t$ (equations \ref{eq:gru} and \ref{eq:pred}), trained on the simulated data described in section \ref{sec:simulation}.

\subsection{Predicting turbulences}
\noindent
Our model predicts the current altitude $d_t$ of the drone using depth images, which is then input to the physical model $\phi$ for the prediction of ground effects \eqref{eq_ground_effect},
\begin{equation}
    f_{a,t} = \phi(d_t, \Omega_t)
\end{equation}
where $\Omega_t$ is the current control input (rotor speeds).

However, the physical parameters of real drones used at deployment may differ from those used in the simulator, requiring identification. For our experiments on the EuRoC-MAV dataset, we used a single flight (\textit{V1\_01\_easy}) for this identification. More precisely, we solve the following optimization problem, using the ground truth heights $d_t$:
\begin{equation}
 \arg \min_{(\alpha, \beta)} \sum_t \left(f_{a,t} - \frac{\alpha \sum_i \Omega_{i,t}^2}{d_t^2-\beta}\right)^2
\end{equation}

\noindent
where $\beta=\mu\frac{r^2}{16}$ and $\alpha=k_T\beta$. We solve the optimization problem using the BFGS algorithm. During testing, we compare the ground truth to ground effects computed as described in \ref{subsec:estimating_gt_turbulence} to the forecasting performed by our model with the identified parameters, that is:
\begin{equation}
\hat f_{a,t} = \frac{\alpha\sum_i \Omega_{i,t}}{\hat d_t^2 - \beta}    
\end{equation}

%\noindent where $\hat d_n$ is obtained via our proposed model from the depth maps.
% %%%%%%%%%%%%%%%%%%%%%%%%%%%%%%%%%%%%%%%%%%%%%%%%%%%%%%%%%%%%%%%%%%%%%%%%%%%%%%%%
% %%%%%%%%%%%%%%%%%%%%%%%%%%%%%%%%%%%%%%%%%%%%%%%%%%%%%%%%%%%%%%%%%%%%%%%%%%%%%%%%
\section{EXPERIMENTS}
\label{sec:experiments}
\noindent
In this section, we present qualitative and quantitative evidence of the performance of our contribution. In detail, we study the effects of adding a temporal memory on the prediction of the turbulences, and the impact of pre-training. Finally, we demonstrate the generalization capabilities of our model on real data.

\subsection{Network architectures}
\noindent
The depth encoder is a 2D convolution network with 6 layers with kernel size of 3 (7 for the first convolution) and output channel size respectively (16, 32, 64, 64, 32, 4), followed by an MLP with one hidden layer (ReLU activation) that project the latent representation into an embedding space $e_t$ of dimension 50. Each convolution is followed by a batch normalization layer and ReLU activation. The decoder used for pre-training has the inverse architecture (an MLP followed by convolutions) coupled with up-sampling layers. The prediction network is a 2-layer GRU with hidden state of size 50, followed by a 2-layers MLP. The final output is a single scalar that represents the height.

\subsection{Training details}
\noindent
We train the transcoder on the IRS dataset and height prediction on the simulated dataset described in section \ref{sec:simulation}. We train on 43 scenes of the train split, optimize network hyper-parameters and perform early stopping on validation scenes, and provide evaluation results on test scenes. In other words, and as common for embodied AI tasks, the prediction performances we report in the tables of this section are given for scenes which have not been seen during training or model validation.

We pre-train the transcoder model with gradient descent and the ADAM update rule and a learning rate of $10^{-3}$ for $500$ epochs and $\lambda$ set to 10,000. We then load and freeze the weights from the encoder head and train the height prediction head (learning rate set to $10^{-6}$) until convergence. Finally, we fine-tune the entire model end-to-end, freeing the parameters of the encoder with the same optimizer configuration. 

\subsection{Experimental setup}
\noindent
The UAVs used during creation of the EuRoC-MAV dataset were equipped with RGB cameras, but not with a depth camera. We circumvented this problem using the  approach in \cite{Gordon2019DepthFV}, which leverages the availability of the 3D scan of the room in which the drone operates. The point cloud of the scene is projected onto the camera plane at each instant of the flight, and the depth images are obtained using simple geometric transformation. Similarly, for evaluation, the ground truth altitude of the drone can be measured using the point cloud. These data were not used during training, and are only seen by the model during the test phase. We focused the evaluation on selected parts of the flights where the ground effect is significant, which mainly consists in take-off and landing phases on the ground or on furniture elevated from the ground and thus recognized as such by the model.

\begin{table}[t!]
\centering
\begin{tabular}{l|l||c|c}
\toprule
\multirow{2}{*}{\textbf{Pre-training}} & \multirow{2}{*}{\textbf{Model}} & \multirow{2}{*}{\textbf{RMSE($d$)}} & \textbf{RMSE($f_a$)}\\
                                       &                                 &                                     & $\times 10^{-4}$\\
\midrule
\multirow{2}{*}{None}                     &\ding{55} GRU & 0.3473 & 39.85 \\ 
                                          &\ding{51} GRU  & 0.3284 & 6.472 \\ \hline
\multirow{2}{*}{Depth$\rightarrow$Depth}  &\ding{55} finetune & 0.3381 & 6.522 \\
                                          &\ding{51} finetune & 0.3283 & 6.428 \\\hline
\multirow{2}{*}{Depth$\rightarrow$Normal} &\ding{55} finetune & 0.3250 & 6.445 \\
                                          &\ding{51} finetune (ours) & 0.3184 & 6.357 \\\bottomrule
\end{tabular}
\caption{\footnotesize{\label{tab:habitat_result}Simulation results on the habitat test set for different neural models: we report the error on height-over-ground estimation (d) and on turbulence ($f_a$).}}
\vspace{-5mm}
\end{table}

\subsection{Experimental results}
\noindent
Estimating ground effect from computer vision is an ill-posed and difficult problem, and our hybrid model achieves good results, as we will show on both the simulated and real data.

\myparagraph{Qualitative results}
Figure \ref{fig:habitat_trajectories} shows example trajectories generated with our procedure described in Section \ref{sec:simulation}. As we can see, the height-prediction is quite accurate even if drone (and thus camera) angles are far from horizontal stable flight paths, which indicates that the model can estimate the correct necessary surface reconstructions.

\myparagraph{Impact of recurrent memory}
The addition of a temporal memory provides a non-negligible gain in altitude prediction, as can be seen in Table \ref{tab:habitat_result}, top rows. Indeed, taking into account the previous depth maps makes it possible not only to keep in memory an invisible surface at frame $t$, but also to smooth the prediction during the hovering phases at constant altitude. Even though the samples requiring memory (because of observation delay) are sparse in the dataset, their impacts are high.
The addition of a GRU is significant especially for low heights where the ground effect $f_a$ is large. The very large difference in $f_a$ (compared to the gain in height estimated $d$) is explained by the fact that ground effect is inverse to square of height and thus error is significantly large for erroneously small predicted heights.

\begin{figure*}[t] \centering
\begin{tabular}{ccc}
\begin{minipage}{0.49\linewidth}
\includegraphics[width=\linewidth]{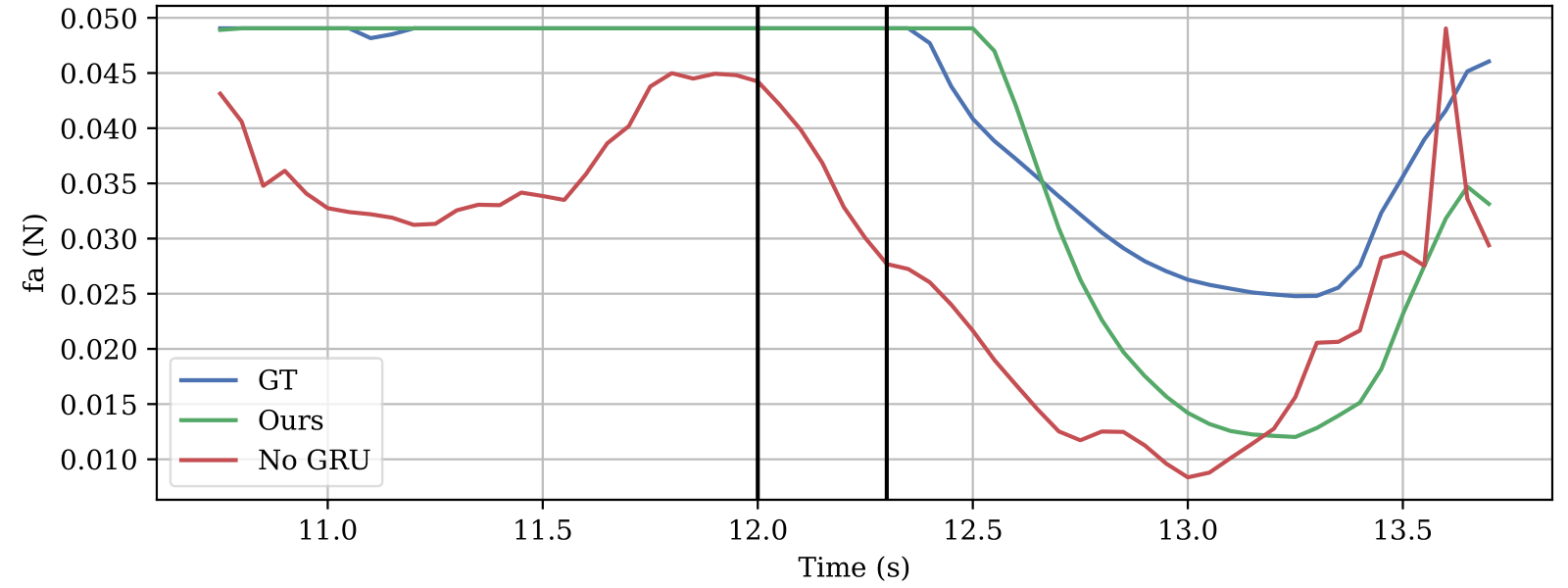}

\end{minipage}
&
\begin{minipage}{0.21\linewidth}
\includegraphics[width=\linewidth]{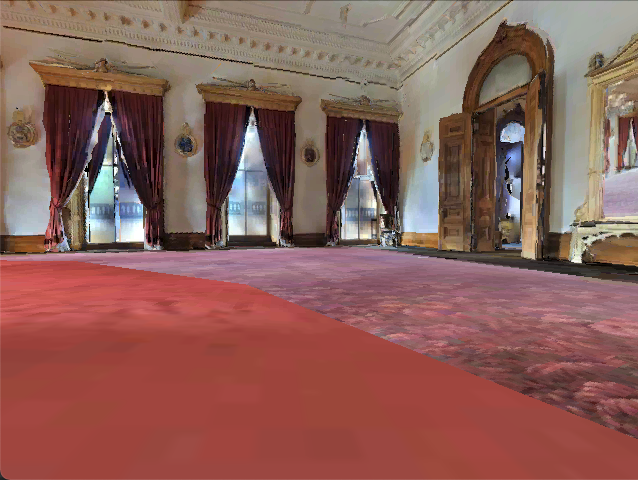}
\vspace{1mm}
\end{minipage}
&
\begin{minipage}{0.21\linewidth}
\includegraphics[width=\linewidth]{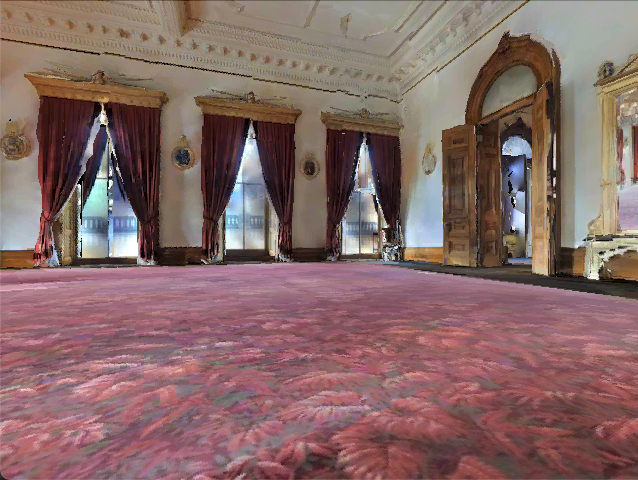}
\vspace{1mm}
\end{minipage}
\\
(a) & (b) & (c) \\
\end{tabular}
\caption{\label{fig:gru}The measurement delay visualized in a single episode: at time $t=12.3$s, indicated as a vertical bar in plot (a), the real height-over-ground is unobserved, as the camera is pointed forward, seen in visual observation (c). The correct height was observed in a previous instant $t=12$, shown in (b).}
\end{figure*}

Figure \ref{fig:gru} illustrates the impact of the recurrent memory on a single episode, where we marked two salient time instants with black vertical bars: at $t=12.3$s, shown in Figure \ref{fig:gru}c, the real height over ground is unobservable, since furniture under the UAV is not visible. This furniture is visible at $t=12$s, shown in Figure \ref{fig:gru}b. The plot in \ref{fig:gru}a shows that this behavior is better captured through the GRU memory of the proposed model.

\begin{table}[t!]
\centering
\begin{tabular}{c|c|c}
\toprule
\textbf{Flight} & \textbf{RMSE($f_{a, \phi}$)} & \textbf{RMSE($f_a$)} \\ 
\midrule
V1 01 (easy)      & 0.1643 & 0.2431 \\ 
V1 02 (medium)    & 0.3127 & 0.5329 \\ 
V3 03 (difficult) & 0.1469 & 0.2822 \\ 
\midrule
Total & 0.2433 & 0.4082 \\ 
\bottomrule
\end{tabular}
\caption{Results on real drone flights of the EuRoC-MAV dataset after training on simulated data only. $f_{a,\phi}$ is the prediction of the identified physics model using GT surface distances $d$. $f_a$ is  the error done by our model accumulating both error of the model $\phi$ and the neural network.}
\label{tab:euroc_result}
\vspace{-5mm}
\end{table}

\begin{figure*}[t] \centering
\includegraphics[width=\linewidth]{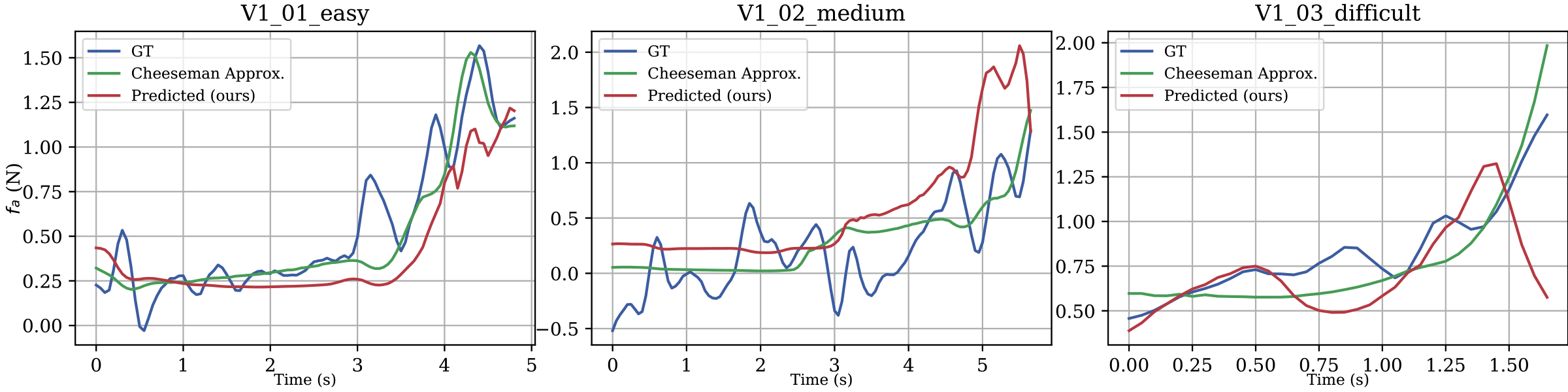}
\caption{\label{fig:sim2realgroundeffect}The gap between simulated and real ground effect on EuRoC-MAV dataset. Blue: actual ground effect as estimated using MoCap data. Red: ground effect as estimated by our hybrid model; Green: ground effect as estimated by the analytic model $\phi$ applied to ground truth heights over ground.}
\vspace{-5mm}
\end{figure*}

\myparagraph{Impact of pre-training}
pre-training the depth encoder is also a key design choice of our method, as can be seen in Table \ref{tab:habitat_result}, middle and bottom rows. Supervising the dense pixel-wise prediction of surface normals facilitates the discovery and reconstruction of planes and thus improves the prediction of height over ground. We also explored an alternative baseline in the form of simply auto-encoding depth images and minimizing reconstruction error, instead of predicting surface normals, indicated as ``Depth$\rightarrow$Depth'' in Table \ref{tab:habitat_result}. While performance is gained w.r.t. to simple end-to-end training from scratch, it is inferior to supervising normals. And unsurprisingly, finetuning the entire network after pre-training provides a gain. We conjecture that fine-tuning helps in focusing extracting the most relevant information (the important ground plane) early on, through the embedding representation $e_t$. With pre-training alone followed using a frozen encoder, the height predictor head must reason on comprehensive representations of the most prominent planes within the image (including vertical and roof planes), since the encoder has not been trained for a specific task.

%this selection must be performed by the height decoder alone from a fixed embedding space $e_t$, which has been trained to represent the full scene and not just the ground plane.

\myparagraph{Sim2real transfer}
The evaluation on the EuRoC-MAV dataset is challenging not only because the depth images constitute out-of-distribution examples for the model, but also because the physical properties of the hexarotor differ strongly from those of the quadrotor used in simulation. Figure \ref{fig:sim2realgroundeffect} shows the evolution of ground effects on episodes from the EuRoC-MAV dataset, and compares the GT turbulences $f_{a,t}$ (blue curve) to two different predictions. The red curve plots the ground effect as estimated by our hybrid model, which first estimates height over ground $d_t$ with a neural network and then turbulence with the analytical ground effect model. To separate the error made by approximating the real ground effect by the analytical model $\phi$, we plot in green the ground effect as estimated by the analytical model applied to ground truth heights over ground $d_t$. While some lower frequency components are lacking, we can see that the analytical model captures the effects quite well.

Table \ref{tab:euroc_result} gives numerical results on the EuRoC-MAV dataset after transfer from simulation. Our model manages to forecast turbulences on the real drone quite precisely. The success of the sim2real transfer can be explained by several phenomena: (i) the use of a hybrid physical and deep learning model brings additional regularization to the network and facilitates extrapolation to new agents, (ii) training on simulation on a wide variety of scenes allows the network to learn to isolate the factors of variation in the scenes.

\myparagraph{Limitations} are shown in figure \ref{fig:sim2realgroundeffect}: although the proposed physical model \eqref{eq_ground_effect} simulates the dominant shape of the turbulence with satisfactory precision, certain non-modeled phenomena produce higher frequency disturbances that $\phi$ does not describe. It is nevertheless possible to adapt the chosen physical law in a more precise form (e.g. blade element theory). However, the prediction of the average value of disturbances at time $t$ already provides a significant amount of useful information to the controller.

\section{Conclusions}
\noindent
We presented a method capable of estimating ground effects generated by a UAV given control inputs and visual observations, from which the model extracts information on the 3D scene structure. We proposed a hybrid model composed of a learned component estimating height over ground from visual data combined with an analytical ground effect model. The neural network model has been designed to take into account negative measurement delays caused by forward facing onboard cameras, and is trained in a large-scale setting from simulated photo-realistic data of 68 different indoor scenes. We also provide a method for pre-training the perception module on additional data with dense supervision of surface normals. While trained in simulation, we show that our method generalizes well to real UAV data from the EuRoC-MAV dataset. We will publish the new simulated dataset as well as the source code of the method upon acceptance.

\bibliographystyle{IEEEtran}
\bibliography{references, references_2}

\end{document}